\documentclass{article}
\usepackage{spconf,amsmath,amssymb,graphicx,booktabs}
\usepackage{color}
\usepackage{enumitem}
\usepackage[hidelinks]{hyperref}
\newcommand{\cots}{\textsc{CoTS}}
\newcommand{\selar}{\textsc{SeLaR}}
\newcommand{\vect}[1]{\mathbf{#1}}

\title{CoTS: Competitive Token Selection for Latent Reasoning in Large Language Models}
\name{Jie Yu$^{1,2}$\sthanks{Work done during an internship at Xiaomi AI Lab.}
\quad Yuxi Sun$^{1}$
\quad Zhaoxu Li$^{2}$
\quad Chunxiao Liu$^{1}$\sthanks{Corresponding authors.}
\quad Xudong Jiang$^{2,\dagger}$}
\address{$^{1}$Xiaomi AI Lab, Beijing, China\\
$^{2}$Nanyang Technological University, Singapore\\
\texttt{jie018@e.ntu.edu.sg}}

\begin{document}
\ninept
\maketitle
\pagestyle{empty}
\thispagestyle{empty}

\begin{abstract}
Recent latent reasoning methods move beyond discrete Chain-of-Thought (CoT) by replacing sampled tokens with continuous soft embeddings or hidden states, enabling richer intermediate representations and preserving alternative reasoning paths. However, the need for latent reasoning varies across decoding steps: confident predictions favor discrete commitment, whereas ambiguous steps benefit from preserving multiple plausible alternatives. Existing approaches typically handle this variation through either global soft activation or hard switching based on uncertainty thresholds, treating latent intervention as an all-or-nothing decision rather than continuously adapting its strength. 
To address this limitation, we propose Competitive Token Selection (CoTS), a training-free framework that reframes latent intervention from binary mode selection into continuous allocation at the token level. CoTS uses the probability competition between the two most likely candidates to dynamically balance discrete commitment and soft exploration at each reasoning step. The second candidate represents the strongest alternative to the model's preferred continuation, so the separation between the two directly reflects the stability of the current discrete choice. When one candidate clearly dominates, CoTS keeps the representation close to the sampled discrete token to preserve a stable reasoning trajectory; as the competition strengthens, it increases the contribution of the soft representation to retain plausible alternatives. This competition based allocation requires neither manually tuned switching thresholds nor mixing coefficients, and introduces no additional forward passes.
Experiments across six reasoning benchmarks demonstrate that CoTS consistently outperforms standard CoT and state-of-the-art training-free latent reasoning methods. Our code is available at \url{https://github.com/OkinoYu/CoTSReasoning}.

\end{abstract}

\begin{keywords}
Large language models, latent reasoning, chain-of-thought, training-free inference, Soft Embeddings
\end{keywords}

\begin{figure*}[t]
    \centering
    \includegraphics[width=0.9\textwidth]{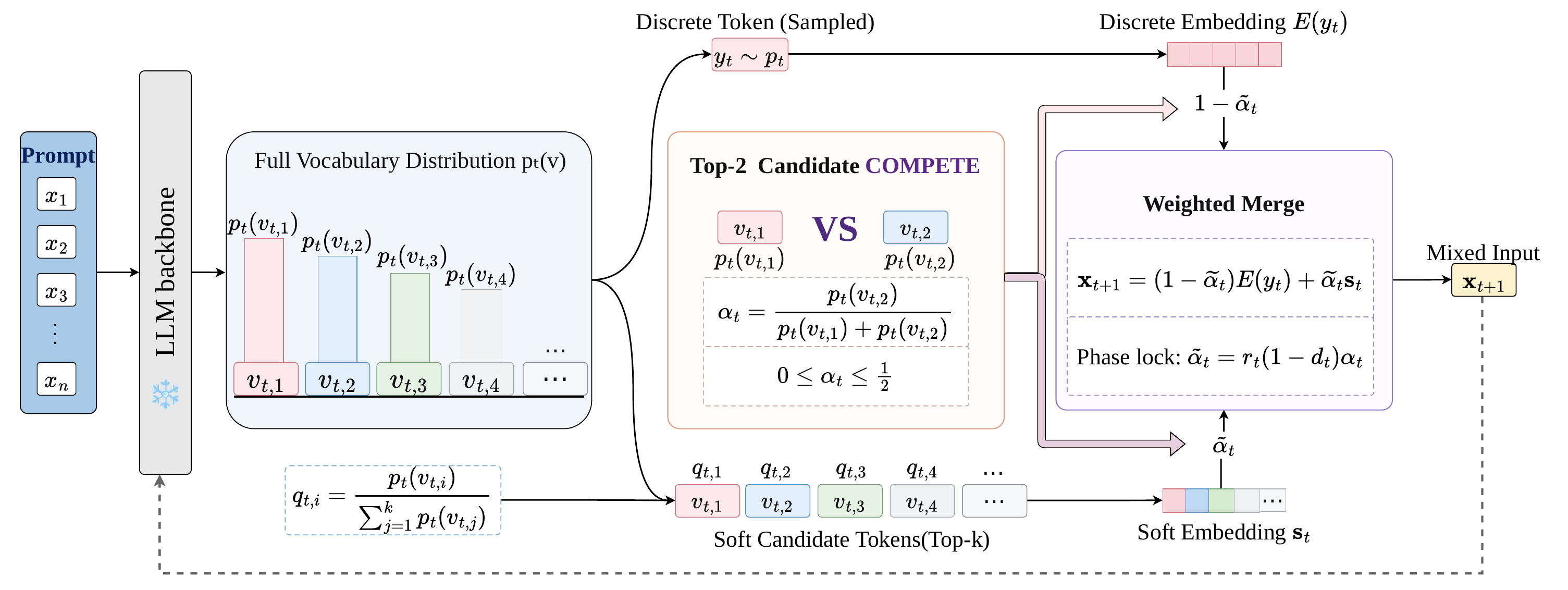}
    \caption{Overview of \cots{}. Top-$k$ candidates form a soft representation, while top-two competition controls its interpolation with the sampled-token embedding, preserving at least half of the weight on the discrete anchor.}
    \label{fig:framework}
\end{figure*}

\section{Introduction}
\label{sec:intro}
Chain-of-thought (CoT) prompting has become a standard paradigm for eliciting multi-step reasoning in large language models (LLMs) \cite{wei2022cot}. Despite its effectiveness, standard autoregressive reasoning proceeds through a sequence of hard token commitments: at each step, the model collapses its predictive distribution into a single discrete token and feeds its embedding back for subsequent generation. While this process produces explicit and interpretable reasoning traces, it discards information carried by alternative plausible continuations and constrains each reasoning step to the discrete language space. Recent latent reasoning methods seek to overcome this limitation by propagating continuous representations, such as hidden states or probability-weighted mixtures of token embeddings, allowing richer information to be retained within each step and multiple plausible reasoning directions to remain represented \cite{zhang2025softthinking,zhuang2025moi,shi2026swireasoning,fu2026selar}. In particular, training-free latent reasoning offers a lightweight way to exploit such continuous representations without modifying the pretrained model parameters.

However, the benefits of latent reasoning are not uniform across decoding steps. Soft representations can preserve alternative reasoning possibilities, but applying them indiscriminately may perturb steps where the model already exhibits a clear preference \cite{zhang2025softthinking,fu2026selar}. This has motivated recent methods to adaptively alternate between discrete and latent reasoning according to model confidence, using entropy-based gating or switching mechanisms to balance exploration and stable convergence \cite{fu2026selar,shi2026swireasoning,li2026later}. Despite their effectiveness, these approaches largely formulate latent intervention as a mode-selection problem: latent reasoning is either activated or suppressed according to a confidence criterion. Yet model uncertainty naturally varies in degree across decoding steps, suggesting that the contribution of latent information should also vary continuously rather than being treated as an all-or-nothing decision. This raises a more fine-grained question: \textit{How much latent information should influence each reasoning step?}

To address this question, we propose \cots{} (Competitive Token Selection), a training-free framework that treats latent reasoning as a continuous allocation between discrete commitment and soft exploration. Rather than measuring uncertainty over the entire predictive distribution, \cots{} focuses on the competition between the two most probable candidates. The second candidate is the strongest alternative to the model's preferred continuation, so their relative probabilities provide a direct signal of how stable the current discrete decision is. \cots{} converts this competition into the amount of latent information introduced at each reasoning step: a decisive prediction keeps the representation close to the sampled discrete token, whereas stronger competition admits more information from plausible alternatives. Importantly, the resulting allocation is naturally bounded so that the discrete token always remains the primary anchor. Latent information therefore acts as a controlled complement to the autoregressive trajectory rather than replacing it. This formulation turns latent intervention from a hard switching decision into a continuous, competition guided allocation, requiring neither manually tuned switching thresholds nor mixing coefficients, and introducing no additional forward passes. Once reasoning is complete, \cots{} returns to standard discrete decoding for final answer generation. Figure~\ref{fig:framework} provides an overview of the framework. Our main contributions are summarized as follows:
\begin{itemize}[leftmargin=5.5mm]

\item We introduce \cots{}, a training-free latent reasoning framework that reframes latent intervention from a binary mode-selection problem into continuous token-level allocation. \cots{} uses top-two probability competition to dynamically balance discrete commitment and soft exploration, without requiring manually tuned switching thresholds or mixing coefficients.

\item We extensively evaluate \cots{} across six reasoning benchmarks and two model families. 
\cots{} consistently improves overall reasoning performance over standard CoT and competitive training-free latent reasoning baselines, achieving 81.48\% average accuracy on Qwen3.5-9B and 86.76\% on Gemma-4-12B-IT, outperforming the strongest latent baseline by 3.44 and 3.74 percentage points, respectively.

\end{itemize}

\begin{table*}[t]
\caption{Pass@1 accuracy (\%) across six reasoning benchmarks. Avg. denotes the unweighted mean across all benchmarks. All methods within each model scale use the same evaluation set and shared sampling settings. method-specific hyperparameters follow their respective implementations. Best and second-best results within each model block are shown in bold and underlined, respectively.}
\label{tab:accuracy}
\centering
\small
\begin{tabular*}{\textwidth}{@{\extracolsep{\fill}}lrrrrrrr@{}}
\toprule
Method & GSM8K & MATH500 & GPQA & AIME24 & AIME25 & AIME26 & Avg. \\
\midrule
\multicolumn{8}{c}{\textit{Gemma-4-12B-IT} \cite{gemmateam2026gemma4}} \\
\midrule
CoT Sampling & \underline{96.82} & \underline{98.40} & \underline{69.19} & 80.00 & \underline{73.33} & 73.33 & 81.85 \\
CoT Greedy & 71.27 & 78.80 & 18.69 & 60.00 & 56.67 & 63.33 & 58.13 \\
Soft Thinking & 82.11 & 86.00 & 46.46 & 80.00 & 70.00 & 56.67 & 70.21 \\
SwiReasoning & 96.66 & 97.60 & 67.17 & \underline{86.67} & \underline{73.33} & \underline{76.67} & \underline{83.02} \\
\selar{} & 94.01 & 91.80 & 57.07 & 80.00 & \textbf{80.00} & 60.00 & 77.15 \\
\cots{} (Ours) & \textbf{97.19} & \textbf{98.60} & \textbf{74.75} & \textbf{93.33} & 70.00 & \textbf{86.67} & \textbf{86.76} \\
\midrule
\multicolumn{8}{c}{\textit{Qwen3.5-4B} \cite{qwen2026qwen35}} \\
\midrule
CoT Sampling & 85.44 & 87.60 & 54.55 & 70.00 & \underline{63.33} & 70.00 & 71.82 \\
CoT Greedy & 75.28 & 71.20 & 23.23 & 53.33 & 33.33 & 30.00 & 47.73 \\
Soft Thinking & \textbf{94.92} & \textbf{94.20} & 54.04 & 70.00 & 60.00 & 70.00 & 73.86 \\
SwiReasoning & \underline{89.08} & \underline{89.80} & \underline{57.07} & \underline{76.67} & \textbf{66.67} & \underline{80.00} & \underline{76.55} \\
\selar{} & 69.60 & 67.20 & 35.86 & 66.67 & 50.00 & 63.33 & 58.78 \\
\cots{} (Ours) & 87.04 & 88.20 & \textbf{61.11} & \textbf{83.33} & \underline{63.33} & \textbf{83.33} & \textbf{77.72} \\
\midrule
\multicolumn{8}{c}{\textit{Qwen3.5-9B} \cite{qwen2026qwen35}} \\
\midrule
CoT Sampling & 89.76 & 92.00 & \textbf{69.19} & 63.33 & 70.00 & \underline{73.33} & 76.27 \\
CoT Greedy & 84.08 & 82.40 & 36.87 & 56.67 & 60.00 & 53.33 & 62.22 \\
Soft Thinking & \textbf{96.36} & \textbf{96.40} & 62.12 & 70.00 & \underline{76.67} & 66.67 & \underline{78.04} \\
SwiReasoning & 89.54 & 88.80 & 60.61 & \underline{76.67} & \textbf{80.00} & 63.33 & 76.49 \\
\selar{} & 77.03 & 69.40 & 44.44 & 56.67 & 70.00 & \underline{73.33} & 65.15 \\
\cots{} (Ours) & \underline{90.83} & \underline{95.20} & \underline{66.16} & \textbf{80.00} & 73.33 & \textbf{83.33} & \textbf{81.48} \\
\bottomrule
\end{tabular*}
\end{table*}

\section{Relation to prior work}
\label{sec:related}
\noindent\textbf{Continuous Latent Reasoning. }While trained approaches like Coconut \cite{hao2025coconut} learn continuous thought states via supervised finetuning, training-free methods manipulate the input space of frozen language models directly \cite{zhang2025softthinking,zhuang2025moi}. Specifically, Soft Thinking \cite{zhang2025softthinking} feeds probability-weighted token embeddings into autoregressive generation, demonstrating that continuous concept representations can compress reasoning steps. Furthermore, recent mechanistic analysis reveals that such soft inputs function as composite continuous prompts rather than branched parallel search trajectories \cite{wu2026singlethreaded}. However, persistent soft feedback can introduce distributional mismatch and representation drift, potentially leading to repetitive or unstable generation
\cite{zhang2025softthinking,shi2026swireasoning,fu2026selar}.

\noindent\textbf{Discrete--Continuous Hybridization. }To curb drift, recent work investigates mechanisms that bridge discrete tokens and continuous representations. Mixture of Inputs (MoI) \cite{zhuang2025moi} formulates an entropy-dependent Bayesian framework to blend discrete tokens with predictive distributions, which requires configuring prior hyperparameters. SwiReasoning \cite{shi2026swireasoning} adopts a macro-level strategy, switching between discrete and latent blocks according to entropy dynamics. At the token level, \selar{} \cite{fu2026selar} applies entropy-gated hard selection and contrastive vector displacement to penalize dominant tokens. In contrast to block-level hard switching \cite{shi2026swireasoning} or ad-hoc vector push-off \cite{fu2026selar}, \cots{} introduces a smooth, step-wise convex interpolation. By governing the soft weight strictly through local top-two probability competition, \cots{} provides an intrinsic 50\% discrete anchor guarantee without tunable thresholds, while preserving purely discrete answer emission via a terminal phase lock.

\section{Competitive Token Selection (\cots{})}
\label{sec:method}

\subsection{Method Overview}
\label{ssec:formulation}
At autoregressive step $t$, a frozen language model predicts the next-token probability distribution $p_t$ over vocabulary $\mathcal{V}$. Standard generation samples a single discrete token $y_t \sim p_t$ and feeds its embedding $E(y_t) \in \mathbb{R}^d$ into the next step. While this guarantees clear and readable text, it immediately discards alternative reasoning possibilities contained in other candidate tokens.

In contrast, training-free soft reasoning methods \cite{zhang2025softthinking} collect the top-$k$ candidate tokens $S_t = \{v_{t,1}, \dots, v_{t,k_t}\} \subseteq \mathcal{V}$ and merge their embeddings into a weighted average vector $\vect s_t$:
\begin{equation}
q_t(v) = \frac{p_t(v)}{\sum_{u \in S_t} p_t(u)},
\qquad
\vect s_t = \sum_{v \in S_t} q_t(v) E(v),
\label{eq:soft}
\end{equation}
where $q_t(v)$ is the normalized probability over the candidate set $S_t$. However, repeatedly feeding soft vectors into a frozen model can introduce distributional mismatch and representation drift, which may result in unstable or repetitive generation
\cite{zhang2025softthinking,shi2026swireasoning,fu2026selar}.

To capture the benefits of both sides—retaining alternative paths while staying grounded in coherent text—\cots{} blends the sampled discrete token embedding with the soft candidate mixture:
\begin{equation}
\vect x_{t+1} = (1 - \widetilde\alpha_t) E(y_t) + \widetilde\alpha_t \vect s_t,
\label{eq:mix}
\end{equation}
where $\widetilde\alpha_t \in [0, 0.5]$ is a dynamic mixing weight that controls the input vector $\vect x_{t+1}$ for the next step.

\subsection{Top-2 Competition Schedule}
\label{ssec:mixing}

A central question is how much latent information should influence each
reasoning step. Rather than summarizing uncertainty over the entire predictive
distribution, we focus on the local competition surrounding the model's
preferred continuation. Let $v_{t,1}$ and $v_{t,2}$ denote the two most
probable candidates, with
$p_t(v_{t,1}) \ge p_t(v_{t,2})$.The runner-up $v_{t,2}$ is the strongest
alternative to the preferred continuation $v_{t,1}$; we therefore define their
competition ratio as $c_t=p_t(v_{t,2})/p_t(v_{t,1})$.

We quantify this competition by the runner-up's normalized probability mass
within the top-two set:
\begin{equation}
\alpha_t =
\frac{c_t}{1+c_t} =
\frac{p_t(v_{t,2})}
{p_t(v_{t,1}) + p_t(v_{t,2})}.
\label{eq:alpha}
\end{equation}

This quantity has a direct allocation interpretation. When the preferred
candidate dominates, $\alpha_t$ approaches zero and the next input remains
close to the sampled discrete token. As the two candidates become comparable,
$\alpha_t$ increases and admits more information from the soft representation.
Since $p_t(v_{t,2}) \le p_t(v_{t,1})$, we have
$0 \le \alpha_t \le 0.5$ without introducing a switching threshold or a
manually chosen mixing coefficient.

Importantly, the top-two candidates determine the \emph{strength} of latent
intervention, whereas the top-$k$ support $S_t$ determines the \emph{content}
of the soft representation. This separation allows \cots{} to use a compact
local competition signal for allocation while retaining richer information
from multiple plausible candidates.

\subsection{Discrete Anchor and Phase Lock}
\label{ssec:guarantee}
\noindent\textbf{Guaranteed Discrete Anchor.}
By expanding Eq.~\eqref{eq:mix}, the mixed input can be written as a
convex combination over $S_t \cup \{y_t\}$. The coefficient assigned to
the sampled token is
\begin{equation}
w_t(y_t)
=
(1-\widetilde{\alpha}_t)
+
\widetilde{\alpha}_t
\mathbf{1}[y_t \in S_t]q_t(y_t)
\ge
1-\widetilde{\alpha}_t
\ge
0.5.
\label{eq:anchor_bound}
\end{equation}
Thus, regardless of whether the sampled token belongs to the latent support, at least half of the total interpolation weight remains explicitly anchored
to the sampled discrete token.

\noindent\textbf{Preventing Representation Drift.}
Because the sampled token always retains at least half of the interpolation weight, the distance between the mixed input $\vect x_{t+1}$ and the actual discrete token embedding $E(y_t)$ is strictly bounded:
\begin{equation}
\|\vect x_{t+1} - E(y_t)\| = \widetilde\alpha_t \|\vect s_t - E(y_t)\| \le 0.5 \|\vect s_t - E(y_t)\|.
\label{eq:bound}
\end{equation}
This bound limits the displacement from the sampled-token embedding to at most half of that induced by the corresponding soft representation, providing a geometric constraint that keeps latent intervention anchored to the discrete reasoning trajectory.

\noindent\textbf{Phase Lock for Final Answers.}
Soft representations are helpful during intermediate reasoning, but the final conclusion and answer formatting require clean, discrete text. To ensure a smooth transition, we introduce two binary control flags:
\begin{equation}
\widetilde\alpha_t = r_t (1 - d_t) \alpha_t,
\label{eq:lock}
\end{equation}
where $r_t \in \{0, 1\}$ indicates whether the model is in the reasoning stage (switching from 1 to 0 immediately once \texttt{</think>} is emitted), and $d_t \in \{0, 1\}$ protects designated math operators or formatting tokens. Once reasoning concludes ($r_t = 0$), $\widetilde\alpha_t$ drops to zero, completely restoring standard discrete token generation.

\noindent\textbf{Computational Efficiency.}
\cots{} is completely training-free and requires only \textit{one} forward pass per generated token. Computing the weighted combination of $k$ candidate embeddings requires only $O(kd)$ basic vector operations ($k \le 9 \ll |\mathcal{V}|$), which adds virtually no latency compared to the model's neural network computation.

\section{Experiments \& Results}
\label{sec:experiments}

\subsection{Experiment Setup}
\label{ssec:setup}

\noindent\textbf{Models.}
We evaluate \cots{} on three reasoning models from two model families:
Qwen3.5-4B and Qwen3.5-9B \cite{qwen2026qwen35}, and
Gemma-4-12B-IT \cite{gemmateam2026gemma4}.
This setting allows us to assess the generality of \cots{} across different model scales and architectures.

\noindent\textbf{Benchmarks.}
We evaluate on six reasoning benchmarks covering grade-school arithmetic, competition-level mathematics, and scientific reasoning: GSM8K (1,319 instances) \cite{cobbe2021gsm8k}, MATH500 (500) \cite{lightman2024verify}, GPQA Diamond (198) \cite{rein2024gpqa}, and AIME 2024--2026 (30 instances each), totaling 2,107 test problems.

\noindent\textbf{Baseline Methods.}
We compare \cots{} against five representative decoding baselines: (1) \textbf{CoT Greedy}, which uses standard greedy argmax decoding; (2) \textbf{CoT Sampling}, which uses standard stochastic autoregressive sampling; (3) \textbf{Soft Thinking} \cite{zhang2025softthinking}, which constructs continuous mixtures of candidate tokens with entropy-based early stopping; (4) \textbf{SwiReasoning} \cite{shi2026swireasoning}, which switches between discrete and latent token blocks according to model confidence; and (5) \textbf{\selar{}} \cite{fu2026selar}, which combines entropy-gated token selection with contrastive vector updates.

\noindent\textbf{Implementation Details.}
We use the official reasoning-enabled chat template for each model backbone. All stochastic methods use random seed 41, temperature 0.6, top-$p=0.95$, repetition penalty 1.0, and a maximum generation length of 38,912 tokens. We set top-$K=20$ for all methods except Soft Thinking, which uses top-$K=30$. For \cots{}, we use $k=9$ on Qwen3.5-9B and $k=3$ on Qwen3.5-4B and Gemma-4-12B-IT, and disable soft mixing after the reasoning stage ends. All experiments are conducted on eight NVIDIA A100 GPUs with per-device batch sizes of 56 for Qwen3.5 and 50 for Gemma-4-12B-IT.


\noindent\textbf{Metrics.}
We report Pass@1 Accuracy (\%) and the unweighted macro-average (Avg.) across all six benchmarks.

\subsection{Results}

As shown in Table~\ref{tab:accuracy}, \cots{} achieves the highest macro-average accuracy on both Qwen3.5-4B and Qwen3.5-9B, reaching 77.72\% and 81.48\%, respectively. Compared with CoT Sampling, \cots{} yields gains of 5.90 and 5.21 percentage points on Qwen3.5-4B and Qwen3.5-9B, respectively. Notably, on Qwen3.5-9B, \cots{} also surpasses the strongest latent reasoning baseline, Soft Thinking \cite{zhang2025softthinking}, by 3.44 percentage points.

The advantage of \cots{} is particularly evident on challenging reasoning benchmarks. Across AIME 2024--2026, \cots{} achieves an average accuracy of 78.89\% on Qwen3.5-9B, outperforming Soft Thinking (71.11\%) and SwiReasoning (73.33\%). On GPQA Diamond, \cots{} reaches 66.16\%, outperforming all evaluated latent reasoning baselines, while Soft Thinking remains stronger on GSM8K and MATH500.

Beyond the Qwen3.5 family, \cots{} achieves the highest average accuracy of 86.76\% on Gemma-4-12B-IT, further improving over CoT Sampling at 81.85\%. This result provides evidence that competition-driven latent allocation transfers across distinct model families and architectures.

\subsection{Ablation Study}

\begin{table}[t]
\centering
\caption{Fixed versus adaptive mixing on Qwen3.5-9B.}
\label{tab:ablation}
\small
\begin{tabular}{lcc}
\toprule
Variant & Mixing weight & Avg. (\%) \\
\midrule
Fixed mixing & $\alpha=0.04225$ & 78.97 \\
\cots{} (Ours) & Adaptive $\alpha_t$ & \textbf{81.48} \\
\bottomrule
\end{tabular}
\end{table}

To verify the core design of \cots{}, we conduct two ablation experiments on Qwen3.5-9B. 

First, we compare our adaptive mixing strategy with a fixed-weight baseline. We set the constant weight to $\alpha=0.04225$, which is the mean $\alpha_t$ observed in the \cots{} run that achieves 81.48\% accuracy. As shown in Table~\ref{tab:ablation}, adaptive competition mixing outperforms the fixed-weight setting by 2.51 percentage points (81.48\% vs.\ 78.97\%). This result indicates that the improvement comes not only from introducing soft information, but also from adjusting its contribution according to the token-level competition at each decoding step.

\begin{table}[!h]
\centering
\caption{Bounded-mapping ablation on Qwen3.5-9B.}
\label{tab:anchor_ablation}
\small
\begin{tabular}{lc}
\toprule
Variant & Avg. (\%) \\
\midrule
w/o Discrete Anchoring ($\alpha_t=c_t$) & 78.51 \\
\cots{} ($\alpha_t=\frac{c_t}{1+c_t}$) & \textbf{81.48} \\
\bottomrule
\end{tabular}
\end{table}

Second, we examine the role of the bounded mixing rule. Replacing our mapping with the unbounded form $\alpha_t=c_t$ allows the soft weight to become as large as 1.0 when the top candidates are equally likely. As shown in Table~\ref{tab:anchor_ablation}, this reduces accuracy from 81.48\% to 78.51\%. The result shows that keeping the sampled discrete token as the dominant input is important for stable mixing.

\subsection{Sensitivity and Stability}
\label{ssec:sensitivity}

\begin{table}[!h]
\centering
\caption{Sensitivity to latent support size $k$ on Qwen3.5-9B.}
\label{tab:topk_sensitivity}
\small
\setlength{\tabcolsep}{3pt}
\begin{tabular}{lccccccc}
\toprule
$k$ & 2 & 3 & 5 & 7 & 9 & 12 & 13 \\
\midrule
Avg. (\%) & 80.63 & 80.67 & 79.97 & 81.44 & \textbf{81.48} & 80.56 & 79.24 \\
\bottomrule
\end{tabular}
\end{table}

\cots{} remains above 80\% across $k=2$--$12$, peaking at $k=9$ with 81.48\%. 
Performance decreases to 79.24\% at $k=13$, suggesting that an overly large support may introduce less informative candidates.

\noindent\textbf{Sampling Stability.}
We further evaluate \cots{} on Gemma-4-12B-IT using four random seeds with all other decoding settings fixed.

\begin{table}[!h]
\centering
\caption{Sampling stability across random seeds.}
\label{tab:seed_stability}
\small
\begin{tabular}{ccccc}
\toprule
Seed & 41 & 42 & 1024 & 2027 \\
\midrule
AIME Avg. (\%) & 83.33 & 81.11 & 82.22 & 82.22 \\
\bottomrule
\end{tabular}
\end{table}

Across the four runs, \cots{} achieves $82.22 \pm 0.91\%$, indicating stable aggregate performance across sampling trajectories.

\section{Conclusion}

We introduced \cots{}, a training-free framework that continuously balances discrete commitment and soft exploration through local token competition. Rather than asking whether reasoning should be discrete or latent, \cots{} asks how strongly a model should commit to its current choice when a plausible alternative still remains. This perspective suggests a broader view of latent reasoning: the boundary between commitment and exploration need not be fixed, but can evolve continuously with the model's own uncertainty.

\clearpage
\nocite{*}
\bibliographystyle{IEEEbib}
\bibliography{refs}

@inproceedings{wei2022cot,
  author = {Jason Wei and Xuezhi Wang and Dale Schuurmans and Maarten Bosma and Brian Ichter and Fei Xia and Ed H. Chi and Quoc V. Le and Denny Zhou},
  title = {Chain-of-Thought Prompting Elicits Reasoning in Large Language Models},
  booktitle = {Advances in Neural Information Processing Systems},
  volume = {35},
  year = {2022},
  url = {https://proceedings.neurips.cc/paper_files/paper/2022/hash/9d5609613524ecf4f15af0f7b31abca4-Abstract-Conference.html}
}

@inproceedings{hao2025coconut,
  author = {Shibo Hao and Sainbayar Sukhbaatar and DiJia Su and Xian Li and Zhiting Hu and Jason Weston and Yuandong Tian},
  title = {Training Large Language Models to Reason in a Continuous Latent Space},
  booktitle = {Proc. Conference on Language Modeling (COLM)},
  year = {2025},
  url = {https://openreview.net/pdf/a7eb656fca52784a118a61fee4aa58ee6ca18e9c.pdf}
}

@inproceedings{zhang2025softthinking,
  author = {Zhen Zhang and Xuehai He and Weixiang Yan and Ao Shen and Chenyang Zhao and Xin Eric Wang},
  title = {Soft Thinking: Unlocking the Reasoning Potential of {LLMs} in Continuous Concept Space},
  booktitle = {Advances in Neural Information Processing Systems},
  volume = {38},
  year = {2025},
  url = {https://proceedings.neurips.cc/paper_files/paper/2025/hash/f7396d1c54d51416958d63e285377103-Abstract-Conference.html}
}

@inproceedings{zhuang2025moi,
  author = {Yufan Zhuang and Liyuan Liu and Chandan Singh and Jingbo Shang and Jianfeng Gao},
  title = {Mixture of Inputs: Text Generation Beyond Discrete Token Sampling},
  booktitle = {Advances in Neural Information Processing Systems},
  volume = {38},
  year = {2025},
  url = {https://proceedings.nips.cc/paper_files/paper/2025/hash/25c61012c845908bff2ea6223cdc6192-Abstract-Conference.html}
}

@inproceedings{shi2026swireasoning,
  author = {Dachuan Shi and Abedelkadir Asi and Keying Li and Xiangchi Yuan and Leyan Pan and Wenke Lee and Wen Xiao},
  title = {{SwiReasoning}: Switch-Thinking in Latent and Explicit for Pareto-Superior Reasoning {LLMs}},
  booktitle = {Proc. International Conference on Learning Representations (ICLR)},
  year = {2026},
  url = {https://proceedings.iclr.cc/paper_files/paper/2026/hash/ddb7bad46132a323aa9d039000941881-Abstract-Conference.html}
}

@inproceedings{fu2026selar,
  author = {Renyu Fu and Guibo Luo},
  title = {{SeLaR}: Selective Latent Reasoning in Large Language Models},
  booktitle = {Proc. 64th Annual Meeting of the Association for Computational Linguistics (Volume 1: Long Papers)},
  year = {2026},
  pages = {7073--7087},
  publisher = {Association for Computational Linguistics},
  doi = {10.18653/v1/2026.acl-long.320},
  url = {https://aclanthology.org/2026.acl-long.320/}
}

@inproceedings{wu2026singlethreaded,
  author = {Junhong Wu and Jinliang Lu and Zixuan Ren and Gangqiang Hu and Zhi Wu and Dai Dai and Hua Wu},
  title = {{LLMs} are Single-threaded Reasoners: Demystifying the Working Mechanism of Soft Thinking},
  booktitle = {Proc. International Conference on Learning Representations (ICLR)},
  year = {2026},
  url = {https://proceedings.iclr.cc/paper_files/paper/2026/hash/0895de150660b6c2aef9b0f098cb1dcd-Abstract-Conference.html}
}

@article{cobbe2021gsm8k,
  author = {Karl Cobbe and Vineet Kosaraju and Mohammad Bavarian and Mark Chen and Heewoo Jun and Lukasz Kaiser and Matthias Plappert and Jerry Tworek and Jacob Hilton and Reiichiro Nakano and Christopher Hesse and John Schulman},
  title = {Training Verifiers to Solve Math Word Problems},
  journal = {arXiv preprint arXiv:2110.14168},
  year = {2021},
  url = {https://arxiv.org/abs/2110.14168}
}

@inproceedings{lightman2024verify,
  author = {Hunter Lightman and Vineet Kosaraju and Yura Burda and Harri Edwards and Bowen Baker and Teddy Lee and Jan Leike and John Schulman and Ilya Sutskever and Karl Cobbe},
  title = {Let's Verify Step by Step},
  booktitle = {Proc. International Conference on Learning Representations (ICLR)},
  year = {2024},
  url = {https://proceedings.iclr.cc/paper_files/paper/2024/file/aca97732e30bcf1303bc22ac3924fd16-Paper-Conference.pdf}
}

@inproceedings{rein2024gpqa,
  author = {David Rein and Betty Li Hou and Asa Cooper Stickland and Jackson Petty and Richard Yuanzhe Pang and Julien Dirani and Julian Michael and Samuel R. Bowman},
  title = {{GPQA}: A Graduate-Level Google-Proof {Q\&A} Benchmark},
  booktitle = {Proc. Conference on Language Modeling (COLM)},
  year = {2024},
  url = {https://openreview.net/pdf?id=Ti67584b98}
}

@inproceedings{shen2025codi,
  author = {Zhenyi Shen and Hanqi Yan and Linhai Zhang and Zhanghao Hu and Yali Du and Yulan He},
  title = {{CODI}: Compressing Chain-of-Thought into Continuous Space via Self-Distillation},
  booktitle = {Proc. Conference on Empirical Methods in Natural Language Processing (EMNLP)},
  year = {2025},
  pages = {677--693},
  publisher = {Association for Computational Linguistics},
  doi = {10.18653/v1/2025.emnlp-main.36},
  url = {https://aclanthology.org/2025.emnlp-main.36/}
}

@inproceedings{wu2025pccot,
  author = {Haoyi Wu and Zhihao Teng and Kewei Tu},
  title = {Parallel Continuous Chain-of-Thought with Jacobi Iteration},
  booktitle = {Proc. Conference on Empirical Methods in Natural Language Processing (EMNLP)},
  year = {2025},
  pages = {914--926},
  publisher = {Association for Computational Linguistics},
  doi = {10.18653/v1/2025.emnlp-main.47},
  url = {https://aclanthology.org/2025.emnlp-main.47/}
}

@inproceedings{su2025tokenassorted,
  author = {Dijia Su and Hanlin Zhu and Yingchen Xu and Jiantao Jiao and Yuandong Tian and Qinqing Zheng},
  title = {Token Assorted: Mixing Latent and Text Tokens for Improved Language Model Reasoning},
  booktitle = {Proc. International Conference on Machine Learning (ICML)},
  volume = {267},
  pages = {57144--57163},
  year = {2025},
  publisher = {PMLR},
  url = {https://proceedings.mlr.press/v267/su25g.html}
}

@inproceedings{gozeten2026cot2,
  author = {Alperen Gozeten and Muhammed Ildiz and Xuechen Zhang and Hrayr Harutyunyan and Ankit Singh Rawat and Samet Oymak},
  title = {Continuous Chain of Thought Enables Parallel Exploration and Reasoning},
  booktitle = {Proc. International Conference on Learning Representations (ICLR)},
  year = {2026},
  url = {https://proceedings.iclr.cc/paper_files/paper/2026/hash/85ec26ef94c3acb4c195e905df1ff4f7-Abstract-Conference.html}
}

@inproceedings{zhou2026lepo,
  author = {Yuyan Zhou and Jiarui Yu and Hande Dong and Zhezheng Hao and Hong Wang and Jianqing Zhang and Qiang Lin},
  title = {{LEPO}: Latent Reasoning Policy Optimization for Large Language Models},
  booktitle = {Findings of the Association for Computational Linguistics: ACL 2026},
  year = {2026},
  pages = {14416--14427},
  publisher = {Association for Computational Linguistics},
  doi = {10.18653/v1/2026.findings-acl.707},
  url = {https://aclanthology.org/2026.findings-acl.707/}
}

@inproceedings{yan2026entrocut,
  author = {Hongxi Yan and Qingjie Liu and Yunhong Wang},
  title = {{EntroCut}: Entropy-Guided Adaptive Truncation for Efficient Chain-of-Thought Reasoning in Small-Scale Large Reasoning Models},
  booktitle = {Proc. IEEE International Conference on Acoustics, Speech and Signal Processing (ICASSP)},
  year = {2026},
  publisher = {IEEE},
  doi = {10.1109/ICASSP55912.2026.11461480},
  url = {https://ieeexplore.ieee.org/document/11461480}
}

@inproceedings{song2026trajectoryentropy,
  author = {Baiyang Song and Zhuoying Feng and Sining Wang and Yi Zhou},
  title = {Beyond Answers: Trajectory Semantic Entropy for Reliable Uncertainty Quantification in {LLMs}},
  booktitle = {Proc. IEEE International Conference on Acoustics, Speech and Signal Processing (ICASSP)},
  year = {2026},
  publisher = {IEEE},
  doi = {10.1109/ICASSP55912.2026.11463008},
  url = {https://ieeexplore.ieee.org/document/11463008}
}

@inproceedings{du2026confidence,
  author = {He Du and Bowen Li and Chengxing Xie and Chang Gao and Kai Chen and Dacheng Tao},
  title = {Exploring Confidence as a Reward to Advance {LLMs} Reasoning},
  booktitle = {Proc. IEEE International Conference on Acoustics, Speech and Signal Processing (ICASSP)},
  pages = {2876--2880},
  year = {2026},
  publisher = {IEEE},
  doi = {10.1109/ICASSP55912.2026.11462883},
  url = {https://ieeexplore.ieee.org/document/11462883}
}

@inproceedings{xu2025softcot,
  author = {Yige Xu and Xu Guo and Zhiwei Zeng and Chunyan Miao},
  title = {{SoftCoT}: Soft Chain-of-Thought for Efficient Reasoning with {LLMs}},
  booktitle = {Proceedings of the 63rd Annual Meeting of the Association for Computational Linguistics (Volume 1: Long Papers)},
  pages = {23336--23351},
  year = {2025},
  publisher = {Association for Computational Linguistics},
  doi = {10.18653/v1/2025.acl-long.1137},
  url = {https://aclanthology.org/2025.acl-long.1137/}
}

@article{li2026later,
  author = {Xuan Li and Yining Wang and Yuchen Liu and Guanjun Liu and Delai Qiu and Shengping Liu and Jiaen Liang and Wei Huang and Jun Yu and Junnan Zhu},
  title = {{LaTER}: Efficient Test-Time Reasoning via Latent Exploration and Explicit Verification},
  journal = {arXiv preprint arXiv:2605.07315},
  year = {2026},
  url = {https://arxiv.org/abs/2605.07315}
}

@misc{qwen2026qwen35,
  author = {{Qwen Team}},
  title = {{Qwen3.5}: Towards Native Multimodal Agents},
  year = {2026},
  howpublished = {Official model release and technical blog},
  url = {https://qwen.ai/blog?id=qwen3.5}
}

@misc{gemmateam2026gemma4,
  author = {{Gemma Team}},
  title = {{Gemma 4 Technical Report}},
  year = {2026},
  eprint = {2607.02770},
  archivePrefix = {arXiv},
  primaryClass = {cs.CL},
  url = {https://arxiv.org/abs/2607.02770}
}

@inproceedings{sun-etal-2026-fact,
  author = {Yuxi Sun and Aoqi Zuo and Haotian Xie and Wei Gao and Mingming Gong and Jing Ma},
  title = {{FACT-E}: Causality-Inspired Evaluation for Trustworthy Chain-of-Thought Reasoning},
  booktitle = {Findings of the Association for Computational Linguistics: ACL 2026},
  pages = {20275--20293},
  year = {2026},
  publisher = {Association for Computational Linguistics},
  doi = {10.18653/v1/2026.findings-acl.1014},
  url = {https://aclanthology.org/2026.findings-acl.1014/}
}

@inproceedings{xiong2026enhancing,
  author = {Siheng Xiong and Ali Payani and Faramarz Fekri},
  title = {Enhancing Language Model Reasoning with Structured Multi-Level Modeling},
  booktitle = {Proc. International Conference on Learning Representations (ICLR)},
  year = {2026},
  url = {https://openreview.net/forum?id=PlkzZhqBCd}
}
\end{document}